\documentclass[10pt,conference,letterpaper]{IEEEtran}
\IEEEoverridecommandlockouts
\usepackage{cite}
\usepackage{amsmath,amssymb,amsfonts}
\usepackage{graphicx}
\usepackage{textcomp}
\usepackage{xcolor}
\usepackage[caption=false,font=normalsize,labelfont=sf,textfont=sf]{subfig}
\usepackage{pifont}
\usepackage{url}
\usepackage[algo2e,linesnumbered,lined,boxed,commentsnumbered,ruled]{algorithm2e}
\usepackage{multirow}
\usepackage{makecell}
\usepackage{relsize}
\usepackage{float}
\usepackage[letterpaper,top=0.75in,left=0.64in,right=0.64in, bottom=1.2in]{geometry}
\def\BibTeX{{\rm B\kern-.05em{\sc i\kern-.025em b}\kern-.08em
    T\kern-.1667em\lower.7ex\hbox{E}\kern-.125emX}}
\begin{document}
% \newgeometry{bottom=4.3cm}
\title{DHO$_2$: Accelerating Distributed Hybrid Order Optimization via Model Parallelism and ADMM}

\author{\IEEEauthorblockN{Shunxian Gu\textsuperscript{*\dag}, Chaoqun You\textsuperscript{\dag}, Bangbang Ren\textsuperscript{*}, Deke Guo\textsuperscript{*}\thanks{Deke Guo is the corresponding author.}, Lailong Luo\textsuperscript{*}, Junxu Xia\textsuperscript{*}}
\IEEEauthorblockA{\textit{\textsuperscript{*}National University of Defense Technology, Changsha, China} \\
\textit{\textsuperscript{\dag}Fudan University, Shanghai, China}\\
% City, Country \\
% email address or ORCID}}
}}
\maketitle

\begin{abstract}
 The surging model and data size boost the development of distributed model training paradigm, which uses a first-order or second-order optimizer as the fundamental tool. FOSI (First-Order and Second-Order Integration), as a hybrid order optimizer, is regarded as a promising substitute due to its fast convergence speed. However, implementing FOSI distributedly will face two challenges: First, the calculation of the curvature information is restricted on a single GPU device, whose memory is unaffordable when the model becomes large and the dimensionality of the curvature information becomes high, hindering the scalability. Second, frequently updating the curvature information incurs high time consumption, which decreases the acceleration of distributed computing. To overcome these challenges, we propose a \textbf{d}istributed \textbf{h}ybrid \textbf{o}rder \textbf{o}ptimization framework, DHO$_2$\footnote{Our source code is available at: \url{https://github.com/ShunxianGu/DHO2}}. It achieves distributed calculation of curvature information via model parallelism to balance the computation and memory cost for each GPU device. Then, it reduces the training time by utilizing the property of ADMM (Alternating Direction Method of Multipliers) on enhancing convergence and proposing an ADMM-like model update rule for the hybrid order optimization setting. Experimentally, our DHO$_2$ can achieve a sublinear time-to-solution and memory usage with the increase of the GPU number, enabling the scalability. Meanwhile, it achieves up to $1.4\times\sim2.0\times$ speedup in the total training time and $4\%\sim5\%$ improvement in the test accuracy, compared with other previous state-of-the-art distributed first- and second-order optimizer frameworks.
% Scaling deep neural network (DNN) training to more devices can reduce time-to-solution. However, it is impractical for users with limited computing resources. FOSI, as a hybrid order optimizer, converges faster than conventional optimizers by taking advantage of both gradient information and curvature information when updating the DNN model. Therefore, it provides a new chance for accelerating DNN training in the resource-constrained setting. In this paper, we explore its distributed design, namely DHO$_2$, 
% including distributed calculation of curvature information and model update with partial curvature information to accelerate DNN training with a low memory burden.
% To further reduce the training time, we design a novel strategy to parallelize the calculation of curvature information and the model update on different devices. 
% Our source code is available at: \url{https://github.com/ShunxianGu/DHO2}.
\end{abstract}


\section{Introduction}
Deep neural network (DNN) models have given rise to numerous productive achievements in research fields such as computer vision \cite{he2016deep,NEURIPS2024_ebc62a3a} and natural language processing \cite{touvron2023llama}. As the size of DNN models and training datasets become larger, training DNN models on a single device is increasingly time-consuming and memory-intensive. For example, finishing a 90-epoch ImageNet-1k training with ResNet-50 (20.5 million parameters) on an NVIDIA M40 GPU takes 14 days \cite{you2018imagenet}. This drawback drives the development of distributed model training which scales single-device training to multiple devices. Usually, the distributed model training has three steps: (\romannumeral1) Replicate the entire model on each device. (\romannumeral2) Each device computes a stochastic gradient on its unique mini-batch training samples in each training iteration. (\romannumeral3) Each device communicates with other devices to synchronize the gradient and update its local model. This procedure repeats
 until the loss function converges and a certain model training
 accuracy is achieved. The whole training process is often referred to as the distributed synchronous stochastic gradient descent (S-SGD) algorithm. S-SGD enables a larger batch size than that of single-device training by dividing training samples into multiple devices. Therefore, it can accelerate model training when confronted with a larger dataset. Such an acceleration paradigm has been widely applied in specific domains, such as autonomous driving \cite{liao2025diffusiondrive} and remote sensing \cite{li2025towards}.

  \begin{figure}[!tbp]
\centering
\subfloat[]{\includegraphics[width=4.4cm]{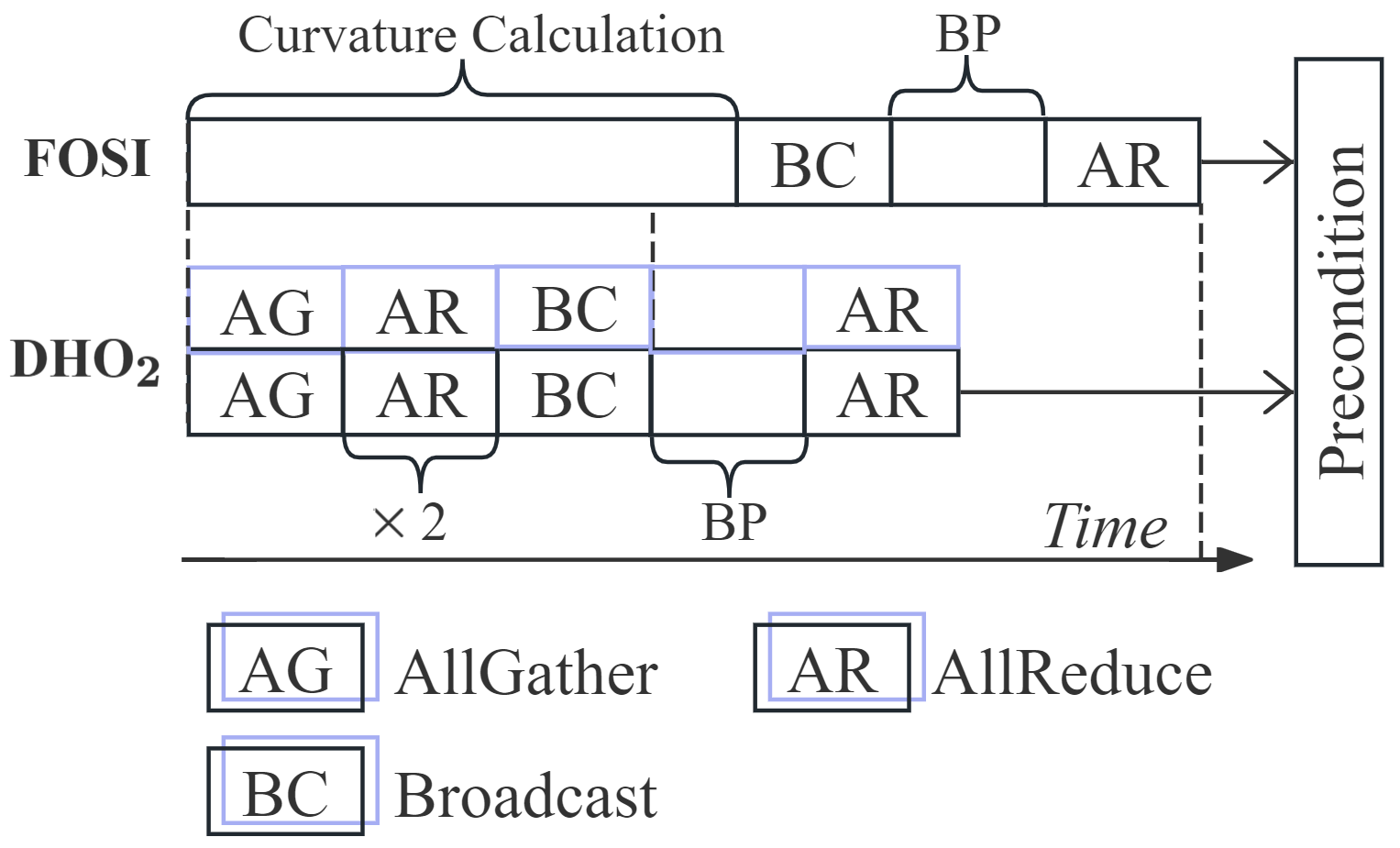}%
} 
% \hfil
\subfloat[]{\includegraphics[width=4.4cm]{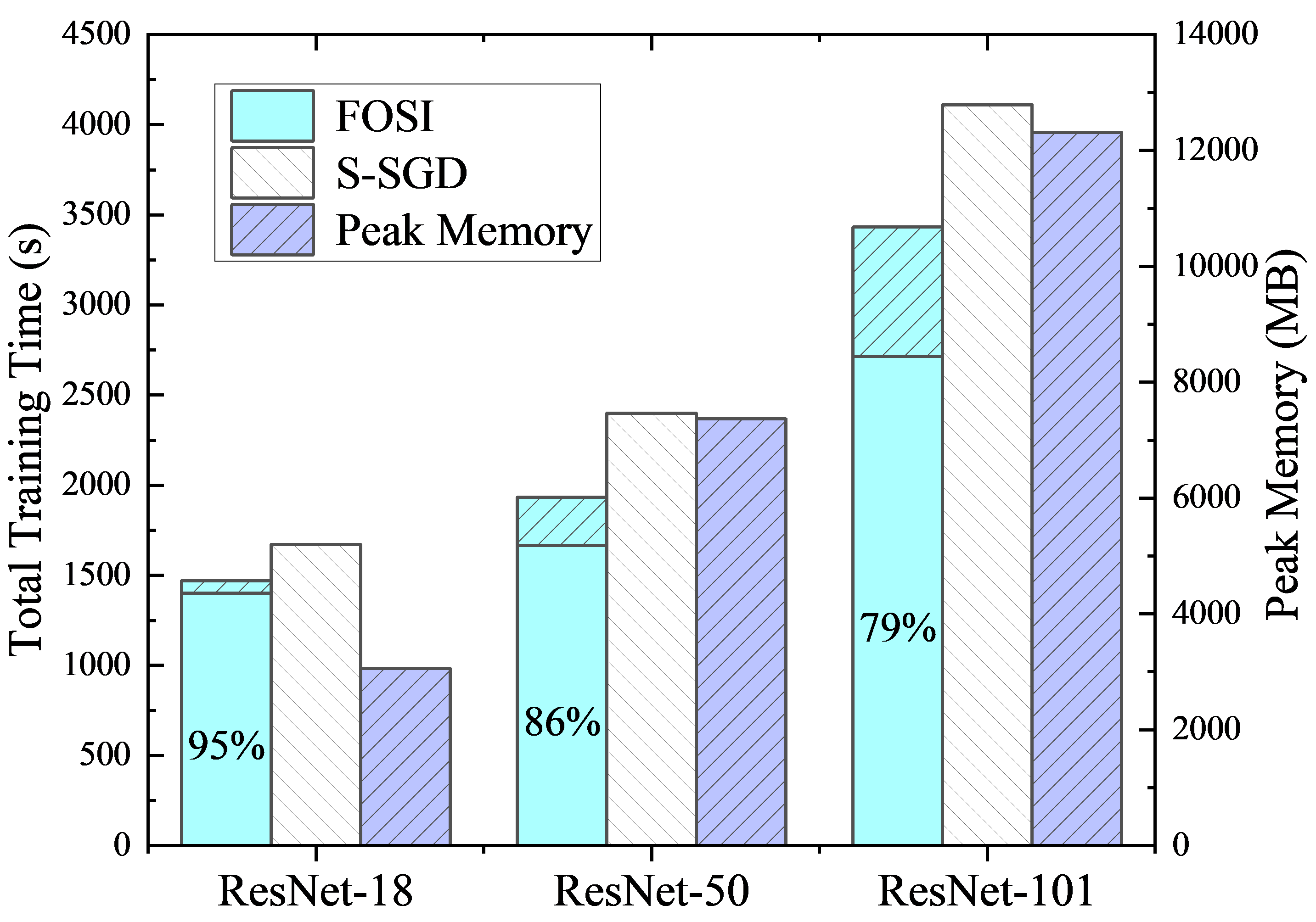}%
}
\caption{Background and motivation. (a) presents the difference between our proposed DHO$_2$ and a simple implementation of FOSI on distributed training, which relies on a single GPU device to calculate the curvature information and broadcast it to the other GPUs for subsequent model training. (b) conducts a toy quick experiment on the ResNet model by comparing the peak memory usage and the total training time between the simple distributed implementation of FOSI and S-SGD on 8 RTX 3090 GPUs. The experimental results demonstrate the challenges described in the paper.}
\label{challenge-a}
\vspace{-0.15cm}
\end{figure}

 Typically, S-SGD utilizes the first-order optimizer (e.g. Adam \cite{kingma2014adam}, RMSProp \cite{hinton2012neural}) and many prior works focus on improving its scalability \cite{bottou2018optimization,you2019large}. However, users with limited financial and computing resources are unable to bear the expense of renting a supercomputer for a long time to run the large-scale S-SGD. Therefore, a natural insight is to replace the optimizer by one with a higher convergence rate to accelerate model training. Recently, second-order optimizers arise since they can get a higher convergence rate than first-order ones by adding curvature information (i.e. Hessian matrix) to model update rules \cite{pauloski2022deep,shi2023distributed}. Due to the high computational burden of calculating curvature information, most of the second-order optimizers \cite{martens2015optimizing,gupta2018shampoo,sivan2022automon} select to approximate it rather than compute the accurate value. This approach sometimes amplifies the approximation error and produces noise, thus degrading the convergence speed. To solve this puzzle, FOSI (First-Order and Second-Order Integration) \cite{sivan2024fosi}, as a hybrid order optimizer, has been proposed by minimizing the loss function with both first- and second-order optimizers. It achieves state-of-the-art model performance and convergence rate when compared with other second-order methods (e.g. K-FAC \cite{martens2015optimizing} and L-BFGS \cite{liu1989limited}).

Although many prior works \cite{gomes2025towards, zhao2024nesterov} have studied the inherent advantages of FOSI, none of them take the distributed model training paradigm into consideration. Implementing FOSI in a distributed training paradigm for further acceleration will face the following two challenges as depicted in the Fig. \ref{challenge-a}: a) FOSI approximates the curvature information through the Lanczos algorithm, which can only be conducted on a single device and its peak memory usage becomes catastrophic when the model becomes large, hindering the scalability. b) FOSI requires frequent updates of the curvature information, which consumes large amounts of training time, leading to a weak advantage on the training efficiency in the distributed setting when compared with the simpler S-SGD algorithm.

To overcome these challenges, we propose DHO$_2$, a scalable FOSI-enabled \textbf{d}istributed DNN training framework based on \textbf{h}ybrid \textbf{o}rder \textbf{o}ptimization. In DHO$_2$, we first design a distributed Lanczos algorithm for curvature information calculation to balance the computation and memory cost for each GPU device via model parallelism \cite{shoeybi2019megatron}. Specifically, by splitting some of the matrix multiplication in the Lanczos algorithm \cite{lanczos1950iteration} into multiple segments and distributing them onto multiple devices, the computational reliance and large memory burden on a single device can be effectively escaped. Next, we propose an ADMM-like model update rule for the hybrid order optimization setting to accelerate model training. Specifically, like other ADMM-based methods \cite{zhou2023federated,bai2025inexact}, we introduce the augmented Lagrangian function and decompose the optimization problem into subproblems to enhance model convergence. However, we further split the subproblem into two orthogonal polynomials and optimize them with first- and second-order optimizers respectively. Such an idea enables DHO$_2$ to make full use of the advantages of ADMM and the hybrid order optimizer on convergence simultaneously.
% Next, we design a distributed model update step to enable each device to update the local model with partial curvature information rather than entire curvature information, which relieves the memory burden effectively. Most importantly, the above two designs resort to the model parallelism technique, which guarantees a computation equivalent to that in single-device training, thereby accelerating model training without degrading model performance. To further accelerate the total time per iteration, we present MR-DHO$_2$, a \textbf{m}ulti-\textbf{r}ole version of DHO$_2$, which classifies devices into two roles, namely accelerators and workers. A parallel algorithm is designed to enable the workers to conduct gradient computations and accelerators to calculate curvature information simultaneously. Moreover, a final model update rule is designed to combine the curvature information and gradient information for further updating the DNN model at the end of each epoch. As depicted in Fig. \ref{lenet} and \ref{MR_DHO2},  compared with DHO$_2$, MR-DHO$_2$ eliminates the training time for the calculation of curvature information on each worker, thereby reducing the average total training time per iteration. 

Our contributions can be summarized as follows: 
\begin{itemize}
    \item We propose a scalable FOSI-enabled distributed hybrid order optimizer framework, namely DHO$_2$, which is also the first distributed implementation of FOSI to the best of our knowledge.
    \item We propose a distributed Lanczos algorithm that allows DHO$_2$ to achieve a faster calculation of curvature information with a lower memory burden than the single-device FOSI theoretically and experimentally.
    \item We propose an ADMM-like model update rule for the hybrid order optimization, that enhances model convergence by making use of the advantages of ADMM and FOSI simultaneously.
    % executes the calculation of curvature information and gradient information in parallel on different devices to further reduce the total training time per iteration.
    \item We conduct an evaluation of DHO$_2$ which achieves up to $1.4\times\sim2.0\times$ speedup to achieve certain model performance compared with other distributed first- and second-order optimizer frameworks.
    \item We conduct a further evaluation of DHO$_2$ about the scalability on a cloud server equipped with up to 32 3090 GPUs and 64 4090 GPUs, which achieves a sublinear time-to-solution and memory usage with the increase of the GPU number.
\end{itemize}

% \vspace{-0.25cm}
\section{Problem Statement and Preliminaries}
In this section, we first introduce some definitions of notations and present our problem statement. Then, we introduce the FOSI optimization method that is originally designed for the single-device training and review its related works.
% \vspace{-0.2cm}
\subsection{Notations and Problem Statement}
For convenience, we unify the notations used in this paper as follows. Non-bold English and Greek letters (e.g. $T,j,\epsilon$ ) denote scalars. Bold lowercase English letters (e.g. $\textbf{g}$) represent vectors while bold uppercase English letters (e.g. $\textbf{H}$) represent matrices. Letters in the calligraphic font (e.g. $\mathcal{D}$) are used to represent sets. Specially, we use $\textbf{diag}(\textbf{v})$ for a diagonal matrix with arbitrary diagonal vector or entries $\textbf{v}$. $\mathbb{O}$ represents the base first-order optimizer. Finally, we define a slice operator, which is a colon ":". For instance, let $\textbf{A}\in\mathbb{R}^{N\times{K}\times{T}}$ denotes a 3-dimensional tensor, $\textbf{A}[n_1:n_2,k_1:k_2,:]$ indicates a sliced sub-tensor from $\textbf{A}$ with $N,K$ indexes ranging from $n_1$ to $n_2$ and $k_1$ to $k_2$ respectively. In particular, in the last dimension $T$, the sub-tensor takes all the entries by marking a single ":". 

The neural network training problem can be posed as a stochastic optimization problem of the form:
% We formulate our constrained acceleration problem as:
\begin{equation}
    \begin{split}
   &\underset{\textbf{w}\in\mathbb{R}^n}{\arg\min}~{\lbrace}f(\textbf{w})=\mathbb{E}_{(\textbf{X},\textbf{Y})\sim\mathcal{D}}=[l(m(\textbf{w,X});\textbf{Y})]{\rbrace} \\
   % \textbf{s.t.}~\frac{1}{C}\sum_{c=1}^Cf(&\textbf{w}_T;\mathcal{D}_c)<\epsilon,~C=G\\
%     &\textbf{X}_l^{(i)T}\textbf{X}_l^{(i)}=\sum_{b=1}^{\lceil|\mathcal{D}_i|/B_l^{(i)}\rceil}{\textbf{X}_l^{\{b\}T}\textbf{X}_l^{\{b\}}} \\
% &\textbf{X}_l^{(i)T}\bar{\textbf{Z}}_l^{(i)}=\sum_{b=1}^{\lceil|\mathcal{D}_i|/B_l^{(i)}\rceil}\textbf{X}_l^{\{b\}T}\bar{\textbf{Z}}_l^{\{b\}}\\
    \end{split}
\label{eq1}
\end{equation}
where $(\textbf{X},\textbf{Y})$ represents a batch of data-label pairs, which is a random sampling from the whole dataset $\mathcal{D}$. $m(\textbf{w,X})$ represents a neural network model that takes as input the model parameters $\textbf{w}$ and the data matrix $\textbf{X}$ and outputs a prediction which has the same dimensionality as $\textbf{Y}$. The loss function $l(;)$ measures how well the prediction matches the target label matrix $\textbf{Y}$. Such an unconstrained non-convex stochastic optimization problem is typically solved via an iterative optimization algorithm. Specifically, given $\textbf{w}_t\in\mathbb{R}^{n}$, the flattened model parameter vector at the $t$-th training iteration, an update step of the form $\textbf{w}_{t+1}=\textbf{w}_t+\textbf{d}_t$ is conducted, where $\textbf{d}_t\in\mathbb{R}^n$ is a descent direction determined by the first- (i.e. gradient $\textbf{g}_t=\nabla{f(\textbf{w}_t)\in\mathbb{R}^n}$) or second-order (i.e. Hessian $\textbf{H}_t=\nabla^2f(\textbf{w}_t)\in\mathbb{R}^{n\times{n}}$) information computed on the current batch of data-label pairs. In the general setting of second-order algorithms, $\textbf{d}_t$ is of the form $-\eta{\textbf{P}}_{t}^{-1}\textbf{g}_t$, where $\textbf{P}_t\in\mathbb{R}^{n\times{n}},~\textbf{P}_t=\textbf{H}_t$(usually) is termed as a preconditioner matrix and $\eta>0$ is a learning rate. Since second-order algorithms take the gradient changes in the future into account, they converge faster than first-order ones. However, computing the full preconditioner matrix is a challenging task and many prior works have made contributions to it. 

% For instance, Shampoo \cite{gupta2018shampoo}, Eva \cite{yang2022efficient} and COMPSO \cite{sun2025compso} exploit the structure of the network to approximate a block diagonal preconditioner matrix, as an alternative to the full one. 
 Newton-Raphson Method \cite{akram2015newton} updates parameters by directly computing and using the inverse of the Hessian matrix, but its high computational cost limits its application in large-scale deep neural networks. To solve this problem, researchers have proposed various improvement strategies, including diagonalization approximation, low-rank decomposition, and Kronecker factorization, to strike a balance between computational efficiency and the fidelity of curvature information. K-FAC \cite{martens2015optimizing} (Kronecker-Factored Approximate Curvature) and Shampoo \cite{gupta2018shampoo}, as representative second-order optimizers based on these strategies, have attracted extensive attention. K-FAC significantly reduces computational complexity by decomposing the Fisher information matrix (FIM) into the Kronecker product, while retaining the key curvature information. Similarly, the Shampoo optimizer constructs layer-wise preconditioners by leveraging the tensor structure of weight matrices and captures the curvature information through a block diagonal preconditioner matrix. Furthermore, based on Shampoo, extended methods such as SOAP \cite{vyas2025soap} and 4-bit Shampoo \cite{wang20244} further combine adaptive learning rate mechanisms or quantization techniques, solving the bottleneck problems of second-order methods in terms of storage and computing resources.

Nonetheless, all the prior works mentioned above approximate the preconditioner matrix directly instead of approximating its inverse, potentially resulting in higher approximation errors and noise sensitivity \cite{li2017preconditioned}. To solve this puzzle, FOSI (First-Order and Second-Order Integration), a hybrid order optimizer, is proposed with three key points: a) It estimates the inverse preconditioner matrix directly, reducing approximation error and computational cost. b) It only estimates the most extreme eigenvalue and vectors of the inverse preconditioner matrix, making it more robust to noise. c) It accepts a base first-order optimizer when it maintains another second-order Newton's optimizer, making it suited for various machine learning tasks. 
% Then, we give an overview of FOSI, which contains four stages: the initialization stage, the warm-up stage, the extreme spectrum estimation (ESE) stage and the model update stage, and then describe the principle in each stage. Notably, our description will be emphasized on the ESE stage and model update stage since they play a key role in our distributed frameworks.

% \begin{equation}
%     \begin{split}
%    &\underset{\textbf{w}_T}{\arg\min}~Tp \\
%    \textbf{s.t.}~\frac{1}{C}\sum_{c=1}^Cf(&\textbf{w}_T;\mathcal{D}_c)<\epsilon,~C=G\\
% %     &\textbf{X}_l^{(i)T}\textbf{X}_l^{(i)}=\sum_{b=1}^{\lceil|\mathcal{D}_i|/B_l^{(i)}\rceil}{\textbf{X}_l^{\{b\}T}\textbf{X}_l^{\{b\}}} \\
% % &\textbf{X}_l^{(i)T}\bar{\textbf{Z}}_l^{(i)}=\sum_{b=1}^{\lceil|\mathcal{D}_i|/B_l^{(i)}\rceil}\textbf{X}_l^{\{b\}T}\bar{\textbf{Z}}_l^{\{b\}}\\
%     \end{split}
% \label{eq1}
% \end{equation}
% where $T$ is the total number of training iterations and $p$ denotes the average total training time per iteration. $C$ is the total number of GPUs (i.e. devices) while $c$ denotes the index of a GPU. $f(\cdot)$, $\textbf{w}_T$, and $\mathcal{D}$ denote the loss function, the final model and data samples respectively. $f(\textbf{w}_T;\mathcal{D}_c)$ represents the loss on the data samples delivered on the $c$-th GPU. $\epsilon$ is the loss bound and $G$ is the limit of the GPU number. The aim of equation \ref{eq1} is to minimize the total training time of our distributed DNN training while a certain model performance is achieved and the computing resources are limited.
% \vspace{-0.2cm}
\subsection{First-Order and Second-Order Integration (FOSI)}\label{s2p2}
FOSI starts by implicitly splitting the optimization problem into two orthogonal subspaces, i.e. $f(\textbf{w})=f_1+f_2$:
\begin{equation}
\begin{split}
f_1=\frac{1}{2}f_t+{(\textbf{w}-\textbf{w}_t)}^T\underbrace{\hat{\textbf{V}}(\hat{\textbf{V}}^T\textbf{g}_t)}_{\textbf{g}_1}+\frac{1}{2}{(\textbf{w}-\textbf{w}_t)}^T\textbf{H}_1{(\textbf{w}-\textbf{w}_t)}\\
f_2=\frac{1}{2}f_t+{(\textbf{w}-\textbf{w}_t)}^T\underbrace{\check{\textbf{V}}(\check{\textbf{V}}^T\textbf{g}_t)}_{\textbf{g}_2}+\frac{1}{2}{(\textbf{w}-\textbf{w}_t)}^T\textbf{H}_2{(\textbf{w}-\textbf{w}_t)}\\
\textbf{H}_1=\hat{\textbf{V}}\textbf{A}_1\hat{\textbf{V}}^T,\textbf{H}_2={\check{\textbf{V}}\textbf{A}_2\check{\textbf{V}}^T},\textbf{A}_1=\textbf{diag}(\hat{\textbf{a}}),\textbf{A}_2=\textbf{diag}(\check{\textbf{a}})
\end{split}
\label{eq3}
\end{equation}
where $\hat{\textbf{a}}\in\mathbb{R}^{k+l}$ denotes a vector which includes the $k$ largest and $l$ smallest eigenvalues of $\textbf{H}_t$, and $\hat{\textbf{V}}\in\mathbb{R}^{n\times{(k+l)}}$ represents a matrix whose columns are the eigenvalues' corresponding eigenvectors. Similarly, we define $\check{\textbf{a}}$ and $\check{\textbf{V}}$ as the vector and matrix including the rest of the eigenvalues and their corresponding eigenvectors respectively. As a result, $\textbf{g}_2$ can be also the form $(\textbf{I}-\hat{\textbf{V}}\hat{\textbf{V}}^T)\textbf{g}_t$ which can be regarded as the Gram-Schmidt orthogonalization of $\textbf{g}_t$ on $\hat{\textbf{V}}$'s column vectors. Therefore, we can infer that $\textbf{g}_2$ is orthogonal to $\textbf{g}_1$ because $\textbf{g}_1$ is the linear combination of the column vectors in $\hat{\textbf{V}}$.

Then, in the arbitrary $t$-th training iteration, FOSI optimizes $f$ by minimizing $f_1$ and $f_2$ independently. For $f_1$, FOSI uses an $\alpha$-scaled Newton’s step. The Netow's step is obtained by putting the derivatives of the first row of the equation \ref{eq3} on $\textbf{w}$ to 0, and is $\alpha$-scaled as follows:

\begin{equation}
    \begin{split}
    &\textbf{w}^*=\textbf{w}_t-\textbf{H}_1^{-1}\textbf{g}_1\\
   \rightarrow\triangle_1&=-\alpha(\hat{\textbf{V}}(\textbf{A}_1^{-1}\hat{(\textbf{V}}^T\textbf{g}_1)))
    \end{split}\label{eq4}
\end{equation}
where $\triangle_1$ denotes the descent vector calculated by Newton's step and it is a linear combination of $\hat{\textbf{V}}$'s column vectors and thus is orthogonal to $\textbf{g}_2$. Meanwhile, FOSI uses the base first-order optimizer to minimize $f_2$. Since certain first-order optimizers (e.g. Adam \cite{kingma2014adam}) can change the direction of $\textbf{g}_2$, to maintain the orthogonality of $\mathbb{O}(\textbf{g}_2)$ to $\triangle_1$, $\mathbb{O}(\textbf{g}_2)$ is further transformed by Gram-Schmidt orthogonalization as follows:
\begin{equation}
\begin{split}
    &\triangle_2=(\textbf{I}-\hat{\textbf{V}}\hat{\textbf{V}}^T)\mathbb{O}(\textbf{g}_2)\\
    =\mathbb{O}(\textbf{g}&_t-\textbf{g}_1)-\hat{\textbf{V}}(\hat{\textbf{V}}^T\mathbb{O}(\textbf{g}_t-\textbf{g}_1))
    \end{split}\label{eq12}
\end{equation}
where $\triangle_2$ denotes the descent vector calculated by the base first-order optimizer. Finally, in the arbitrary $t$-th training iteration, the model update step is mathematically represented as $\textbf{w}_{t+1}=\textbf{w}_t+\triangle_1+\triangle_2$.

% \begin{figure}[H]
%     \centering
% \begin{algorithm2e}
% \caption{Lanczos algorithm}
% \LinesNumbered
% \SetKwData{Left}{left}\SetKwData{This}{this}\SetKwData{Up}{up}
% \SetKwFunction{Union}{Union}\SetKwFunction{FindCompress}{FindCompress}
% \SetKwInOut{Input}{Input}\SetKwInOut{Output}{Output}
% \Input{$m,hvp_t$;}
% \Output{$\textbf{D},\textbf{B}$;}
% \BlankLine
% initialization: initialize $\textbf{D}$ and $\textbf{B}$ with zero matrices; \\
% initialization: randomize the column vector $\textbf{D}[:,1]$ with standard Gaussian distribution and normalize it to a unit vector;\\
% \For{$i=1~to~m$}{
% $\textbf{v}_i\leftarrow~\textbf{D}[:,i]$;\\
% $\textbf{h}\leftarrow~hvp_t(\textbf{v}_i)$;\\
% $\textbf{B}[i,i]\leftarrow~\textbf{h}\centerdot{\textbf{v}_i}$\\
% /* Gram-Schmidt orthogonalization */\\
% $\textbf{h}\leftarrow~\textbf{h}-\textbf{D}(\textbf{D}^T\textbf{h})$\\
% /* Gram-Schmidt orthogonalization */\\
% $\beta\leftarrow~\Vert{\textbf{h}}\Vert_F$\\
% $\textbf{B}[i+1,i]\leftarrow~\beta$,~$\textbf{B}[i,i+1]\leftarrow~\beta$\\
% $\textbf{D}[:,i+1]\leftarrow~\textbf{h}/\beta$\\
% }
% % \setlength{\belowdisplayskip}{-10pt}
% \end{algorithm2e}
%     % \caption{Caption}
%     % \label{fig:enter-label}
% \end{figure}
\begin{algorithm2e}[!tb]
\caption{Lanczos algorithm}\label{al1}
\LinesNumbered
\SetKwData{Left}{left}\SetKwData{This}{this}\SetKwData{Up}{up}
\SetKwFunction{Union}{Union}\SetKwFunction{FindCompress}{FindCompress}
\SetKwInOut{Input}{Input}\SetKwInOut{Output}{Output}
\Input{$m,hvp_t$;}
\Output{$\textbf{D},\textbf{B}$;}
\BlankLine
initialization: initialize $\textbf{D}$ and $\textbf{B}$ with zero matrices; \\
initialization: randomize the column vector $\textbf{D}[:,1]$ with standard Gaussian distribution and normalize it to a unit vector;\\
\For{$i=1~to~m$}{
$\textbf{v}_i\leftarrow~\textbf{D}[:,i]$;\\
$\textbf{h}\leftarrow~hvp_t(\textbf{v}_i)$;\\
$\textbf{B}[i,i]\leftarrow~\textbf{h}\centerdot{\textbf{v}_i}$\\
/* Gram-Schmidt orthogonalization */\\
$\textbf{h}\leftarrow~\textbf{h}-\textbf{D}(\textbf{D}^T\textbf{h})$\\
/* Gram-Schmidt orthogonalization */\\
$\beta\leftarrow~\Vert{\textbf{h}}\Vert_F$\\
$\textbf{B}[i+1,i]\leftarrow~\beta$,~$\textbf{B}[i,i+1]\leftarrow~\beta$\\
$\textbf{D}[:,i+1]\leftarrow~\textbf{h}/\beta$\\
}
\end{algorithm2e}
% \DecMargin{5em}
% \vspace{-0.7cm}

According to the model update step, FOSI should only calculate the stochastic gradient $\textbf{g}_t$ via BP algorithm in each training iteration while the eigendecomposition of $\textbf{H}_1$ needs to be calculated and updated once every a pre-set number (i.e. $I$) of training iterations. Such a model update step allows FOSI to estimate the most extreme eigenvalues and vectors of the Hessian matrix instead of approximating the full one, reducing large computational cost. Moreover, FOSI estimates the eigendecomposition of $\textbf{H}_1$ and its inverse directly via the Lanczos algorithm. Such a way enables FOSI to obtain a full low-rank representation of the Hessian for the first subspace $\hat{\textbf{V}}$, which captures both the rotation and curvature of the sub-problem $f_1$, contributing to the accuracy and stability of the optimization. The pseudo-code of the Lanczos algorithm is shown in algorithm \ref{al1}. Specifically, the Lanczos algorithm takes the number of Lanczos iterations $m=max\{4(k+l),2\ln{n}\}$ and an operator $hvp_t(\textbf{v})=\textbf{H}_t\textbf{v},~\forall{\textbf{v}\in\mathbb{R}^n}$ as the input. The operator is generated by the Pearlmutter’s algorithm \cite{pearlmutter1994fast} using the loss function $f(\cdot)$ and the latest model parameter $\textbf{w}_t$. In the initialization step (line 1$\sim$2 of the algorithm \ref{al1}), $\textbf{B}\in\mathbb{R}^{m\times{m}}$ and $\textbf{D}\in\mathbb{R}^{n\times{m}}$ are initialized by creating zero matrices with the same shapes. They are ultimately a tridiagonal matrix and an orthogonal matrix (in which each column vector is a unit vector and orthogonal to each other) respectively. $\textbf{D}[:,1]$, which is the first column vector of $\textbf{D}$, is initialized by a standard normal distribution and then normalized to a unit vector. In the arbitrary $i$-th Lanczos iteration, $\textbf{B}[i,i]$ should be calculated to satisfy the following equation with $\textbf{v}_i$, which is the $i$-th column vector of $\textbf{D}$, i.e. $\textbf{D}[:,i]$.
\begin{equation}
\textbf{B}[i,i]=\textbf{v}_i^T\underbrace{\textbf{H}_t\textbf{v}_i}_{hvp_t(\textbf{v}_i)}
\label{eq5}
\end{equation}
This equation is realized in lines 4$\sim$6. Then, the Lanczos algorithm computes a vector $\textbf{h}$ orthogonal to $\textbf{v}_i$ in lines 7$\sim$9, normalizes it to a unit vector in line 12 and serves it as $\textbf{D}[:,i+1]=\textbf{v}_{i+1}$. Meanwhile, in line 10$\sim$11, $\textbf{B}[i+1,i]$ and $\textbf{B}[i,i+1]$ are calculated to satisfy the following equations:

\begin{equation}
\begin{cases}
    \textbf{B}[i+1,i]=\textbf{v}_{i+1}^T\textbf{H}_t\textbf{v}_i{=}\Vert{\textbf{h}}\Vert_F\\
    \textbf{B}[i,i+1]=\textbf{v}_{i}^T\textbf{H}_t\textbf{v}_{i+1}{=}\Vert{\textbf{h}}\Vert_F
\end{cases}
% \textbf{v}_i^T\underbrace{\textbf{H}_t\textbf{v}_i}_{hvp_t(\textbf{v}_i)}=\textbf{B}[i,i]
\label{eq6}
\end{equation}
 After the Lanczos iterations finish, the resultant $\textbf{D}$ and $\textbf{B}$ satisfy the following equation:
\begin{equation}
\begin{split}
    &\textbf{D}^T\textbf{H}_t\textbf{D}=\textbf{B}\\
    \Leftrightarrow \underbrace{\textbf{D}\textbf{D}^T}_{\textbf{I}}&\textbf{H}_t\underbrace{\textbf{D}\textbf{D}^T}_{\textbf{I}}=\textbf{D}\textbf{B}\textbf{D}^T\\
    \Leftrightarrow&\textbf{H}_t=\textbf{D}\textbf{B}\textbf{D}^T
\end{split}
% \textbf{v}_i^T\underbrace{\textbf{H}_t\textbf{v}_i}_{hvp_t(\textbf{v}_i)}=\textbf{B}[i,i]
\label{eq7}
\end{equation}
since $\textbf{D}$ is an orthogonal matrix whose columns are formed by $\{\textbf{v}_i\}_{i=1}^m$. By taking the eigendecomposition of
$\textbf{B}$, we can get the following equation:
\begin{equation}
\textbf{H}_t=\textbf{DU}\textbf{diag}(\textbf{u})\textbf{U}^T\textbf{D}^T\Leftrightarrow\textbf{H}_t=\underbrace{(\textbf{DU})}_{\textbf{Z}}\textbf{diag}(\textbf{u})\underbrace{(\textbf{DU})}_{\textbf{Z}}{}^T
\label{eq8}
\end{equation}
where $\textbf{u}$ is the vector formed by the eigenvalues of $\textbf{B}$ and $\textbf{U}$ is the matrix whose columns are the corresponding eigenvectors of the eigenvalues. Since not all the eigenvalues in $\textbf{u}$ are the approximation of the eigenvalues in $\textbf{H}_t$, the equation \ref{eq8} is an approximate eigendecomposition of $\textbf{H}_t$. However, the most extreme values in $\textbf{u}$ can converge to the real most extreme eigenvalues of $\textbf{H}_t$ with the increase of $m$. Finally, we can get the required $\hat{\textbf{a}}$ and $\hat{\textbf{V}}$ in the equation \ref{eq3} by extracting the $k$ largest and $l$ smallest entries from $\textbf{u}$ and selecting their corresponding eigenvectors from $\textbf{Z}$. $\textbf{H}_1^{-1}$ is thus obtained.

  \begin{figure}[!tb]
\centerline{\includegraphics[width=9cm]{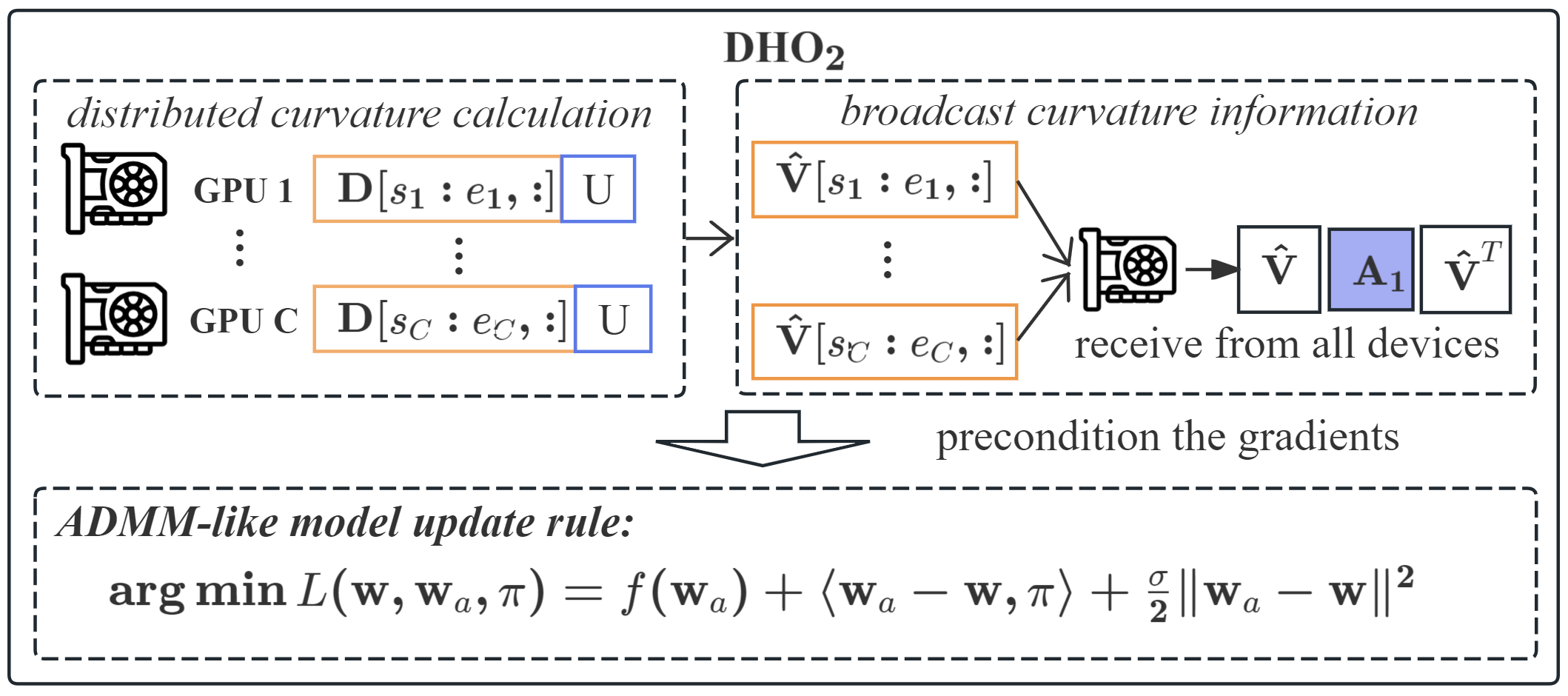}}
\caption{An overview of DHO$_2$.}
\label{DHO2_overview}
% \vspace{-0.8cm}
\vspace{-0.3cm}
\end{figure}

\section{Framework Design}
 In this section, we introduce the proposed DHO$_2$ framework, which encompasses a distributed curvature information calculation algorithm and an ADMM-like model update rule for the hybrid order optimization setting. An overview of DHO$_2$ is shown in Fig. \ref{DHO2_overview}.

\subsection{Distributed Lanczos Algorithm}\label{p3p1}
As the model becomes deeper and wider, the number of model parameters $n$ becomes extremely large. As a result, it is unaffordable for the memory space of a single GPU to store $\textbf{D}\in\mathbb{R}^{n\times{m}}$. Therefore, the target of the distributed Lanczos algorithm is to remain a part of $\textbf{D}$ on each GPU before each GPU achieves the eigendecomposition of $\textbf{H}_1$ synchronously. The pseudo-code of the distributed Lanczos algorithm is shown in algorithm \ref{al2}, where $s_c$ and $e_c$ are the split start point and the split end point on the $c$-th GPU. In general, $e_c-s_c+1=n/C$, where $C$ is the total GPU number. Specifically, different from the original Lanczos algorithm, each GPU only initializes its part of $\textbf{D}$ in line 1 and fixes the first column vector partially in line 3. Then, in the arbitrary $i$-th Lanczos iteration, each GPU gathers parts of $i$-th column vectors from the other GPUs to form the complete $\textbf{v}_i$ so that $\textbf{B}[i,i]$ can be updated by using the equation \ref{eq5} (in line $5$$\sim$line $7$). The model parallelism technique is introduced in the following Gram-Schmidt orthogonalization (line 8$\sim$ line 11) to obtain a correct but partial $\textbf{h}[s_c:e_c]$. At the end of the Lanczos iteration, since each GPU stores a part of $\textbf{h}$, the Frobenius norm of $\textbf{h}$ should be obtained by calculating the $l_2$ norm of $\textbf{h}[s_c:e_c]$ on each GPU and synchronizing it with an \textit{all-reduce} operation. The square root of the synchronization result is the required Frobenius norm, as in lines 12$\sim$13. The next column vector of $\textbf{D}$ which is partially stored on each GPU is updated by the normalized $\textbf{h}[s_c:e_c]$, as in line 14. After finishing the Lanczos iterations, the distributed Lanczos algorithm obtains the same $\textbf{B}$ as the original Lanczos algorithm and the partial $\textbf{D}[s_c:e_c,:]$. Next, we decompose the $\textbf{B}$ to $\textbf{U}\textbf{diag}(\textbf{u})\textbf{U}^T$ via the eigendecomposition and thus we can get the $\hat{\textbf{a}}$ and partial $\hat{\textbf{V}}[s_c:e_c,:]\in\mathbb{R}^{n/C\times(k+l)}$ by extracting the $k$ largest and $l$ smallest entries from $\textbf{u}$ and selecting its corresponding partial column vectors from $\textbf{D}[s_c:e_c,:]\textbf{U}$. Finally, by broadcasting each GPU's own partial $\hat{\textbf{V}}[s_c:e_c,:]$ to the others, we can get the complete $\hat{\textbf{V}}$ required for the calculation of $\textbf{H}_1^{-1}$ and its inverse. Thanks to the model parallelism technique, each GPU can afford partial computation and memory burden and the computational result is identical to that of the original one.

% \textbf{MEM-OPT:} As shown in Fig. \ref{DHO2_overview}, we also propose a memory-optimized version in which the partial $\hat{\textbf{V}}[s_c:e_c,:]$ on each GPU is further split into pieces and scattered to other GPUs, instead of the broadcast operation. Such a way enables

% Based on the model parallelism, we outline the distributed Lanczos algorithm (), which is executed in parallel on multiple devices. Obviously, Different from the original work, we longitudinally partition the $U\in{R^{m\times{n}}}$ onto $D$ devices, where $D$ is the total number of devices.
% \vspace{-0.2cm}

\begin{algorithm2e}[!tb]
\caption{distributed Lanczos algorithm}\label{al2}
\LinesNumbered
\SetKwData{Left}{left}\SetKwData{This}{this}\SetKwData{Up}{up}
\SetKwFunction{Union}{Union}\SetKwFunction{FindCompress}{FindCompress}
\SetKwInOut{Input}{Input}\SetKwInOut{Output}{Output}
\Input{$m,hvp_t$;}
\Output{$\textbf{D}[s_c:e_c,:],\textbf{B}$;}
\BlankLine
initialization: initialize $\textbf{D}[s_c:e_c,:]$ and $\textbf{B}$ with zero matrices; randomize $\textbf{v}_1$ by standard Gaussian distribution with a same random seed and normalize it to a unit vector;\\
\For{$c=1~to~C$ \textbf{in parallel}
}{
$\textbf{D}[s_c:e_c,1]$=$\textbf{v}_1[s_c:e_c]$\\
\For{$i=1~to~m$}{
$\textbf{v}_i\leftarrow~all\_gather(\{\textbf{D}[s_c:e_c,i]\}_{c=1}^C)$;\\
$\textbf{h}\leftarrow~hvp_t(\textbf{v}_i)$;\\
$\textbf{B}[i,i]\leftarrow~\textbf{h}\centerdot{\textbf{v}_i}$\\
/* Gram-Schmidt orthogonalization */\\
$\textbf{m}\leftarrow{all\_reduce}(\{\textbf{D}[s_c:e_c,:]^T\textbf{h}[s_c:e_c]\}_{c=1}^C)$\\
$\textbf{h}[s_c:e_c]\leftarrow~\textbf{h}[s_c:e_c]-\textbf{D}[s_c:e_c,:]\textbf{m}$\\
/* Gram-Schmidt orthogonalization */\\
$\beta\leftarrow~\sqrt{all\_reduce(\{\Vert\textbf{h}[s_c:e_c]\Vert_2^{2}\}_{c=1}^C)}$\\
$\textbf{B}[i+1,i]\leftarrow~\beta$,~$\textbf{B}[i,i+1]\leftarrow~\beta$\\
$\textbf{D}[s_c:e_c,i+1]\leftarrow~\textbf{h}[s_c:e_c]/\beta$\\
}
}
\end{algorithm2e}

\subsection{ADMM-like Model Update Rule}\label{s3p2}
Before we propose our ADMM-like model update rule, we first rewrite the optimization problem in the equation \ref{eq1} by introducing an auxiliary variable, $\textbf{w}_a=\textbf{w}$, as follows:
\begin{equation}
\underset{\textbf{w}_a,\textbf{w}\in\mathbb{R}^n}{\arg\min}~f(\textbf{w}_a)~\textbf{s.t.}~\textbf{w}_a=\textbf{w}\\
    \label{eq9}
\end{equation}

ADMM has proven its effectiveness in enhancing model convergence \cite{ebrahimi2024aa, zhou2025preconditioned}. The backgrounds of ADMM can be referred to the earliest work \cite{gabay1976dual} and a nice book \cite{boyd2011distributed}. To apply ADMM for the equation \ref{eq9}, we introduce the augmented Lagrange function defined by,
\begin{equation}
L(\textbf{w},\textbf{w}_a,\pi)=f(\textbf{w}_a)+\langle{\textbf{w}_a-\textbf{w},\pi}\rangle+\frac{\sigma}{2}\|{\textbf{w}_a-\textbf{w}}\|^2\\
    \label{eq10}
\end{equation}
where $\pi\in\mathbb{R}^n$ is the Lagrange Multiplier and $\sigma>0$. The framework of ADMM for the problem \ref{eq9} is given as follows: for an initialized point ($(\textbf{w}_a^0,\textbf{w}^0,\pi^0)$), performing the following updates iteratively for every $k\geq0$,

\begin{equation}
\begin{cases}
    \textbf{w}^{k+1}=\underset{\textbf{w}\in\mathbb{R}^n}{\arg\min}~L(\textbf{w},\textbf{w}_a^k,\pi^k)=\textbf{w}_a^k+\frac{\pi^k}{\sigma}\\
    \textbf{w}_a^{k+1}=\underset{\textbf{w}_a\in\mathbb{R}^n}{\arg\min}~L(\textbf{w}^k,\textbf{w}_a,\pi^k)\\
    \pi^{k+1}=\pi^k+\sigma(\textbf{w}_a^{k+1}-\textbf{w}^{k+1})
\end{cases}
% \textbf{v}_i^T\underbrace{\textbf{H}_t\textbf{v}_i}_{hvp_t(\textbf{v}_i)}=\textbf{B}[i,i]
\label{eq11}
\end{equation}

We intend to utilize the hybrid order optimization to accelerate the solving of the subproblem 2 in the equation \ref{eq11}, by splitting the Lagrange function \ref{eq10} into two orthogonal subspaces, $L(\textbf{w},\textbf{w}_a,\pi)=L_1+L_2$,
\begin{equation}
\begin{split}
L_1=\frac{1}{2}f_t+{(\textbf{w}-\textbf{w}_a^t)}^T{\textbf{g}_1}&+\frac{1}{2}{(\textbf{w}-\textbf{w}_a^t)}^T\textbf{H}_1{(\textbf{w}-\textbf{w}_a^t)}\\
+\langle{\textbf{w}_a-\textbf{w},\hat{\textbf{V}}\hat{\textbf{V}}^T\pi}&\rangle+\frac{\sigma}{2}\|{\textbf{w}_a-\textbf{w}}\|^2\\
L_2=\frac{1}{2}f_t+{(\textbf{w}-\textbf{w}_a^t)}^T{\textbf{g}_2}&+\frac{1}{2}{(\textbf{w}-\textbf{w}_a^t)}^T\textbf{H}_2{(\textbf{w}-\textbf{w}_a^t)}\\
+\langle{\textbf{w}_a-\textbf{w},\check{\textbf{V}}\check{\textbf{V}}^T\pi}&\rangle
% \textbf{H}_1=\hat{\textbf{V}}\textbf{A}_1\hat{\textbf{V}}^T,\textbf{H}_2={\check{\textbf{V}}\textbf{A}_2\check{\textbf{V}}^T},\textbf{A}_1=\textbf{diag}(\hat{\textbf{a}}),\textbf{A}_2=\textbf{diag}(\check{\textbf{a}})
\end{split}
\label{eq13}
\end{equation}
Then, we can use the $\alpha$-scaled Newton's optimizer and the base first-order optimizer to optimize the first and second equations of \ref{eq13} respectively,
\begin{equation}
\begin{split}
\triangle_1&=-\alpha(\hat{\textbf{V}}((\textbf{A}_1+\sigma\textbf{I})^{-1}\hat{(\textbf{V}}^T(\hat{\textbf{V}}\hat{\textbf{V}}^T(\textbf{g}_t+\pi))))))\\
\triangle_2&=\mathbb{O}(\textbf{g}_t+\pi-\textbf{g}_1)-\hat{\textbf{V}}(\hat{\textbf{V}}^T\mathbb{O}(\textbf{g}_t+\pi-\textbf{g}_1))
\end{split}
\label{eq14}
\end{equation}
Since the second condition of \ref{eq11} requires that the $\textbf{w}_a$ should converge to the stationary point which satisfy ${\nabla}f(\textbf{w}_a)+\pi=0$, the update of $\textbf{w}_a^{k+1}$ should experience the following inner loop until satisfying the following condition,
\begin{equation}
\begin{split}
&\textbf{w}_a^{l+1}=\textbf{w}_a^{l}+\triangle_1+\triangle_2,\textbf{w}_a^0=\textbf{w}^k\\
until~&\|\textbf{g}_l+\pi+\sigma(\textbf{g}_l+\pi)\|_2^2<\epsilon,\textbf{w}_a^{k+1}=\textbf{w}_a
\end{split}
\label{eq15}
\end{equation}
where $\epsilon$ is a very small number. In the practical implementation, such a condition is satisfied by a number of repeated updates of $\textbf{w}_a^l$ which is conducted by repeatedly computing $\triangle_1$ and $\triangle_2$ on the whole dataset. The details of the DHO$_2$ are shown in algorithm \ref{al3}, where $P$ is the number of inner loops. 
In the rest of the paper, $KP$ denotes the number of total training epochs.

\begin{algorithm2e}[!tb]
\caption{DHO$_2$}\label{al3}
\LinesNumbered
\SetKwData{Left}{left}\SetKwData{This}{this}\SetKwData{Up}{up}
\SetKwFunction{Union}{Union}\SetKwFunction{FindCompress}{FindCompress}
\SetKwInOut{Input}{Input}\SetKwInOut{Output}{Output}
% \Input{$m,hvp_t$;}
% \Output{$\textbf{D}[s_c:e_c,:],\textbf{B}$;}
\BlankLine
initialize $\alpha,\eta,\sigma,K,P>0$;\\
initialize $\pi^0=0,\textbf{w}_a^0=\textbf{w}^0$;\\
\For{$k=0~to~K$
}{
% $\textbf{D}[s_c:e_c,1]$=$\textbf{v}_1[s_c:e_c]$\\
$\hat{\textbf{a}}$,$\hat{\textbf{V}}$ $\leftarrow~distributed\_lanczos\_algorithm()$  \\
$\textbf{w}^{k+1}=\textbf{w}_a^k+\frac{\pi^k}{\sigma}$\\
$\textbf{w}_a^{0}=\textbf{w}$\\
\For{$l=0~to~P$}{
\For{$(\textbf{X,Y})~$sampled from the Dataset $\mathcal{D}$}{
$\textbf{g}_t:$each GPU computes the stochastic gradient and synchronizes it via an \textit{all-reduce} operation\\
$\textbf{w}_a^{l+1}=\textbf{w}_a^{l}+\triangle_1+\triangle_2$\\
}
% $\hat{\textbf{a}}$,$\hat{\textbf{V}}$ $\leftarrow~distributed\_lanczos\_algorithm()$  \\
% $\textbf{v}_i\leftarrow~all\_gather(\{\textbf{D}[s_c:e_c,i]\}_{c=1}^C)$;\\
% $\textbf{h}\leftarrow~hvp_t(\textbf{v}_i)$;\\
% $\textbf{B}[i,i]\leftarrow~\textbf{h}\centerdot{\textbf{v}_i}$\\
% /* Gram-Schmidt orthogonalization */\\
% $\textbf{m}\leftarrow{all\_reduce}(\{\textbf{D}[s_c:e_c,:]^T\textbf{h}[s_c:e_c]\}_{c=1}^C)$\\
% $\textbf{h}[s_c:e_c]\leftarrow~\textbf{h}[s_c:e_c]-\textbf{D}[s_c:e_c,:]\textbf{m}$\\
% /* Gram-Schmidt orthogonalization */\\
% $\beta\leftarrow~\sqrt{all\_reduce(\{\Vert\textbf{h}[s_c:e_c]\Vert_2^{2}\}_{c=1}^C)}$\\
% $\textbf{B}[i+1,i]\leftarrow~\beta$,~$\textbf{B}[i,i+1]\leftarrow~\beta$\\
% $\textbf{D}[s_c:e_c,i+1]\leftarrow~\textbf{h}[s_c:e_c]/\beta$\\
}
$\textbf{w}_a^{k+1}=\textbf{w}_a$\\
$\pi^{k+1}=\pi^k+\sigma(\textbf{w}_a^{k+1}-\textbf{w}^{k+1})$\\
}
\Output{$\textbf{w}_a$}
\end{algorithm2e}
\addtolength{\topmargin}{0.2cm}

\subsection{Computation, Memory and Communication Analysis}\label{p3p4}
% Before we start the complexity analysis of DHO$_2$ with the original FOSI, we first denote the number of epochs as $E$ and assume that each epoch contains $I$ training iterations. In other words, DHO$_2$ and the original FOSI start the ESE stage at the beginning of each epoch. Meanwhile, since both DHO$_2$ and the original FOSI compute stochastic gradients in their model update step, it is unnecessary to take the computational complexity of the gradient computation into consideration.
\textbf{\textit{Computation}}: The disparity between DHO$_2$ and the original FOSI resides in the incorporation of model parallelism in the Gram-Schmidt orthogonalization process. In the Gram-Schmidt orthogonalization of the original FOSI, the computation in line 8 of algorithm \ref{al1} requires a complexity of $\mathcal{O}(2mn+n)$. Therefore, the total reduction of the computational complexity from FOSI to DHO$_2$ on the curvature information for each GPU is $\mathcal{O}((2m+1)(n-n/C))$. Moreover, although DHO$_2$ introduces extra computations (line 5, line 13, line 14 of the algorithm \ref{al3}), it is trivial when compared with the acceleration brought by enhanced convergence, which will be proved in Section \ref{p5p2}.

\textbf{\textit{Memory:}} As mentioned in Section \ref{p3p1}, the key idea of alleviating the memory burden of FOSI is to reduce the memory usage of $\textbf{D}$, which occupies the largest memory usage compared with other objects. In the Lanczos algorithm of the original FOSI, the main source of the memory burden comes from the variables $\textbf{D,B,v,h}$. Thus, the complexity of the peak memory usage becomes $\mathcal{O}(mn+m^2+n)$ on a single device. However, in DHO$_2$, each GPU needs to store a part of $\textbf{D,h}$. Therefore, the peak memory usage of each GPU is reduced to $\mathcal{O}(mn/C+m^2+n/C)$.

\textit{\textbf{Communication:}} DHO$_2$ introduces four communication operations (one \textit{all-gather}, two \textit{all-reduce} and one \textit{broadcast}) in the distributed Lanczos algorithm while it requires one \textit{all-reduce} operation at the beginning of each epoch, just as S-SGD. We do think such a sacrifice on the communication is worthy for a scalable implementation. Furthermore, such a sacrifice can be alleviated by the enhanced convergence, which can reduce the training epochs required for achieving certain model performance, resulting in performing less times of the distributed Lanczos algorithm.

\section{Experiments}
\subsection{Experimental Settings}
\textbf{Testbed.} We implement our DHO$_2$ framework on a cloud server which is equipped with 32 NVIDIA®GeForce®RTX 3090 GPUs and 64 NVIDIA®GeForce®RTX 4090 GPUs. Each GPU is linked by RoCE with a bandwidth of 50 Gbps.

\begin{table*}[!tbp]
\renewcommand{\arraystretch}{1.3}
\setlength\tabcolsep{8.0pt}
\caption{Table of the frameworks' required training epochs to achieve a certain accuracy and total training time to achieve their optimal model performance.}
\vspace{-0.65cm}
% \vspace{-0.7cm}
\begin{center}
\centering
\subfloat{
\begin{tabular}{c|c|ccc|c}
%添加顶部横线 
\Xhline{1.5 pt}
%输入标题
\multirow{2}{*}{\textbf{Frameworks}} & \multirow{2}{*}{\textbf{Models}} & \multicolumn{3}{c|}
{\textbf{Accuracy}} & \multirow{2}{*}{\textbf{Time(s)}}\\
\cline{3-5}
&&$81\%$&$86\%$&$90\%$&\\
\cline{1-6}
DHO$_2$&\multirow{4}{*}{VGG-16}&16&35&\textbf{83}&\textbf{3265}\\
\cline{3-5}
DHO$_2$-WA&&23&41&\textbf{91}&\textbf{3579}\\
\cline{3-5}
D-KFAC&&23&36&\textbf{97}&\textbf{5044}\\
\cline{3-5}
D-Shampoo&&10&66&INF&INF\\

\Xhline{1.5 pt}
\end{tabular}}
\subfloat{
\begin{tabular}{c|c|ccc|c}
%添加顶部横线 
\Xhline{1.5 pt}
%输入标题
\multirow{2}{*}{\textbf{Frameworks}} & \multirow{2}{*}{\textbf{Models}} & \multicolumn{3}{c|}
{\textbf{Accuracy}} & \multirow{2}{*}{\textbf{Time(s)}}\\
\cline{3-5}
&&$62\%$&$66\%$&$67\%$&\\
\cline{1-6}
DHO$_2$&\multirow{4}{*}{ResNet-101}&20&\textbf{74}&97&\textbf{1004}\\
\cline{3-5}
DHO$_2$-WA&&29&\textbf{77}&INF&\textbf{1047}\\
\cline{3-5}
D-KFAC&&26&\textbf{89}&INF&\textbf{3560}\\
\cline{3-5}
D-Shampoo&&99&INF&INF&INF\\

\Xhline{1.5 pt}
\end{tabular}}
\end{center}
\vspace{-0.8cm}
\label{tab1}

\end{table*}

\textbf{Models and dataset.} To evaluate our framework's effectiveness, we test it on the CIFAR-10/100 datasets \cite{krizhevsky2009learning} by training VGG-16 \cite{simonyan2014very} and ResNet-101 \cite{he2016deep} models respectively on these datasets. Both of the CIFAR-10 and CIFAR-100 datasets contain 60,000 tiny images which have a resolution of $32\times32$. The difference is that the images in the CIFAR-10 dataset can be classified into 10 categories while those in the CIFAR-100 dataset can be classified into 100 categories. Furthermore, to prove the scalability of our framework, we train the ResNet-152 \cite{he2016deep} model on the tiny-imagenet dataset \cite{tiny-imagenet}, which contains 200 classes and each class has 500 images for training, with two different types of GPUs respectively. All images in the tiny-imagenet dataset are 64x64 RGB ones.

\textbf{Baselines and metrics.} To evaluate the speedup of our framework, we compare it with three baselines and record their required training time to achieve a certain accuracy as the main metric. The three baselines are chosen since they are once the state-of-the-art frameworks that use gradient preconditioning to accelerate model training and are shown as follows:

\begin{itemize}
    \item \textbf{DHO$_2$-WA:} A version of DHO$_2$ without the ADMM-like model update rule, to prove the effectiveness of ADMM on enhancing convergence.
    \item \textbf{Distributed K-FAC} \cite{pauloski2022deep}: A well-known second-order distributed model training framework that preconditioned the gradients by Fisher Information matrix.
    \item \textbf{Distributed shampoo} \cite{shi2023distributed}: A novel PyTorch distributed implementation of the shampoo algorithm which preconditions the gradients via a diagonal preconditioner matrix.
\end{itemize}

Then, we compare the time-to-solution of DHO$_2$ with that of S-SGD when our testbed is equipped with different numbers of GPUs to prove the scalability. The time-to-solution is the total training time to achieve the optimal accuracy on the tiny-imagenet dataset for S-SGD/DHO$_2$ within a pre-set number of training epochs. Furthermore, we record the peak memory usage of each GPU device for DHO$_2$ to prove the effectiveness of the distributed Lanczos algorithm on the memory burden.
% is utilized as the main metric while the total training time per iteration and memory cost are also taken into account. Specifically, in our first experiment, we compare DHO$_2$ with the original FOSI to evaluate the effectiveness of our distributed design on accelerating training time and alleviating memory burden, which is discussed in the section \ref{p3p4}. In our second experiment, we compare DHO$_2$ and MR-DHO$_2$ with distributed first-order (Adam \cite{kingma2014adam} and RMSProp \cite{hinton2012neural}) and second-order (K-FAC \cite{pauloski2022deep}) optimizer frameworks to prove the superiority of our frameworks on the training time. By the way, regardless of whether our experiment is conducted in a single-device or multi-device setting, the total batch size is consistently set to 256 since this value can achieve the best convergence speed empirically. In MR-DHO$_2$, $50\%$ of the GPUs are treated as the accelerators since this ratio can eliminate the bubble in the parallelization at large.

\textbf{Hyperparameter setting:} For our DHO$_2$, we select the best combination of hyperparameters empirically. Thus, the $\sigma$ is chosen as 5e-4, 5e-6, 5e-7 for the ResNet-101, VGG-16 and ResNet-152 models respectively. $P$ is set to 4. Since our intention is to prove the speedup of our framework when achieving competitive model performance as in \cite{garg2021dct,qu2024fedqclip}, we choose the total training epochs as 100 and thus $K=25$ to ensure our models converge to such model performance. For all frameworks aforementioned, we utilize the AdamW \cite{loshchilov2017decoupled} optimizer as the base first-order optimizer with a learning rate of $\eta=1e$-3 and a weight decay of $0.05$ for a fair comparison.

\subsection{Comparison with Second-Order Frameworks}\label{p5p2}
\begin{figure}[!tbp]
\centering
\subfloat[CIFAR-10]{\includegraphics[width=4.4cm]{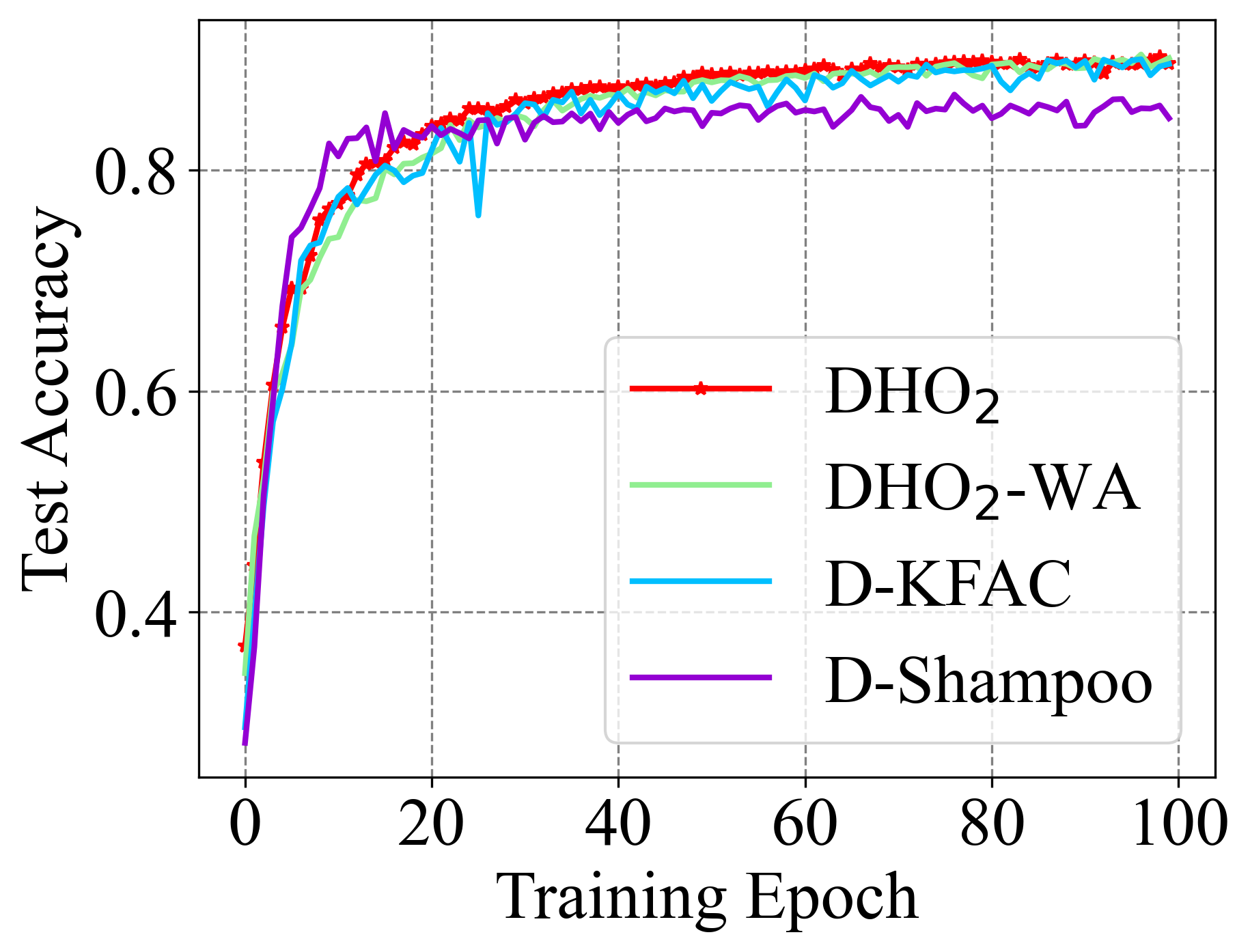}%
} 
% \hfil
\subfloat[CIFAR-100]{\includegraphics[width=4.4cm]{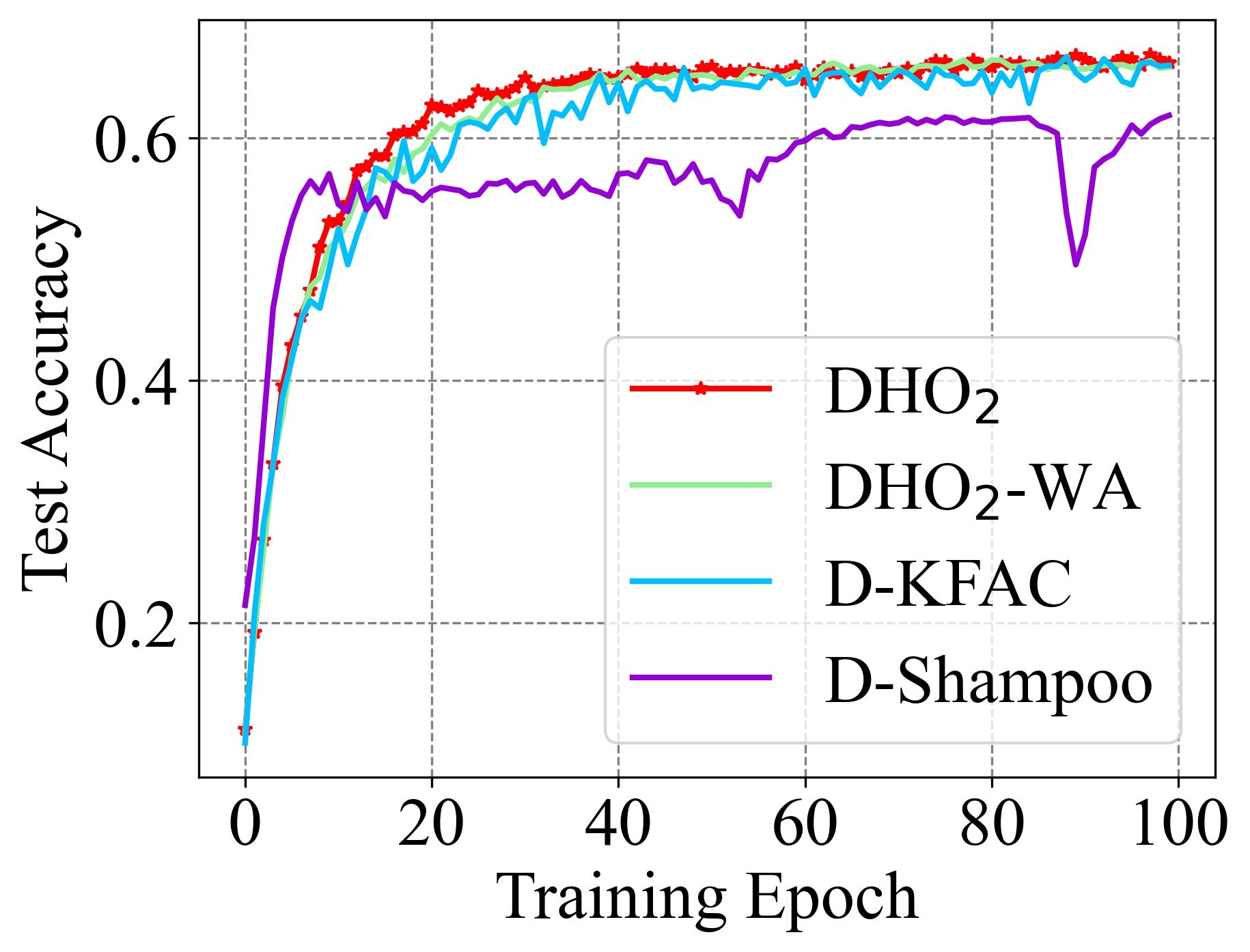}%
}
\caption{The test accuracy versus training epoch curves on the CIFAR-10/100 datasets. DHO$_2$ achieves the highest test accuracy ($90\%/67\%$) within 100 training epochs and requires the fewest training epochs to reach it.}
\label{ex1}
\vspace{-0.3cm}
\end{figure}

We conduct our first experiment on 16 3090 GPUs with a fixed batch size of 16 on each GPU. Fig. \ref{ex1} and Table. \ref{tab1} demonstrate the test accuracy versus training epoch curve and some checkpoints of the curve. We also record the total training time in the table. However, since DHO$_2$-WA, D-KFAC and D-Shampoo cannot converge to the optimal accuracy that DHO$_2$ achieves on the CIFAR-100 dataset, we record the total training time they require to converge to the suboptimal accuracy (66\%) instead. Table. \ref{tab1} shows that DHO$_2$ can reduce $5\%\sim10\%$ training time and converge to a better test accuracy when compared with DHO$_2$-WA, proving the effectiveness of our ADMM-like model update rule. Moreover, both DHO$_2$ and DHO$_2$-WA converge $1.5\times\sim2\times$ faster than D-KFAC, thanks to the fast convergence of their fundamental hybrid order optimizer. Meanwhile, D-Shampoo cannot converge to such performance as DHO$_2$ and D-KFAC, though it requires a greatly shorter training time (4.72 s/epoch and 5.63 s/epoch on the CIFAR-10 and CIFAR-100 datasets respectively). Thus, we do think our DHO$_2$ achieves the state-of-the-art performance both in the training time and test accuracy, compared with the second-order optimizer frameworks.
% In order to prove that DHO$_2$ has a lower computational complexity and memory burden than the original FOSI, we implement them on training DenseNet-201. Fig. \ref{ex1} demonstrates the test accuracy versus training epoch curve, the total training time per epoch and the peak memory usage of DHO$_2$ and the original FOSI. The curve shows that DHO$_2$ requires $8\sim10$ training epochs to achieve 90\% accuracy while the original FOSI needs 12 epochs. The total training epochs of them are close. That is because the introduction of the model parallelism technique enables DHO$_2$ to have the same computation process as the original FOSI. As a result, DHO$_2$ has the same convergence rate as the original FOSI. Meanwhile, DHO$_2$ has a shorter total training time per epoch than the original FOSI, verifying our computational complexity analysis in the section \ref{p3p4}. However, with the increase of the GPU number, the total training time per epoch does not exhibit a linear downward trend. This is because the introduction of more GPUs can result in an extra communication burden.
% Finally, the peak memory usage exhibits an approximate linear reduction since each GPU can store a smaller piece of the matrix $\textbf{D}$ with the increase of GPU number, as discussed in the section \ref{p3p4}.

% \begin{figure}[!tbp]
% \centerline{\includegraphics[width=9cm]{fig_ex1.png}}
% \caption{The model performance, total training time per epoch and peak memory usage of DHO$_2$ and the original FOSI.}
% \label{ex1}
% \vspace{-0.4cm}
% \end{figure}

  \begin{figure}[!tbp]
\centering
\subfloat[24 3090 GPUs]{\includegraphics[width=4.4cm]{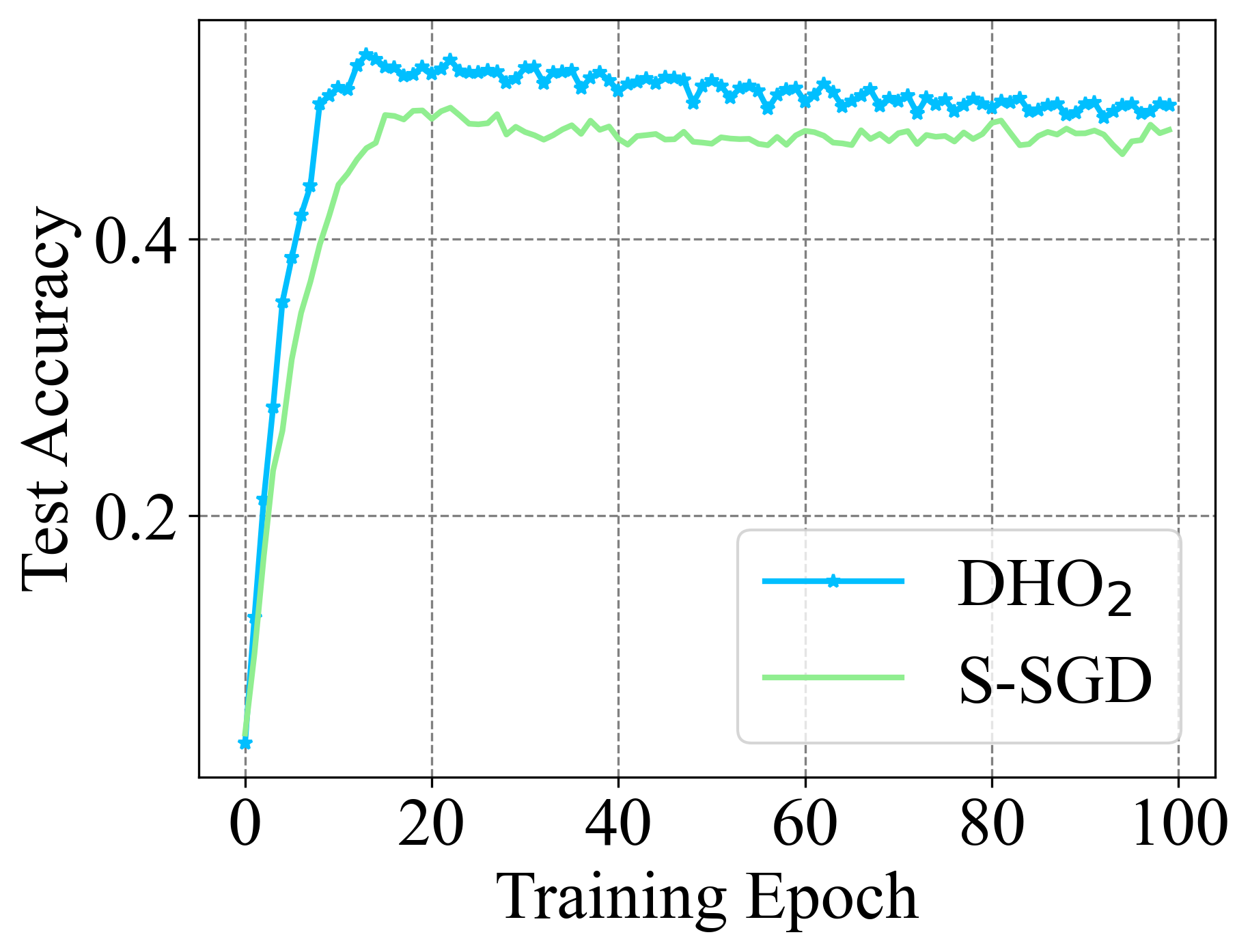}%
} 
% \hfil
\subfloat[64 4090 GPUs]{\includegraphics[width=4.4cm]{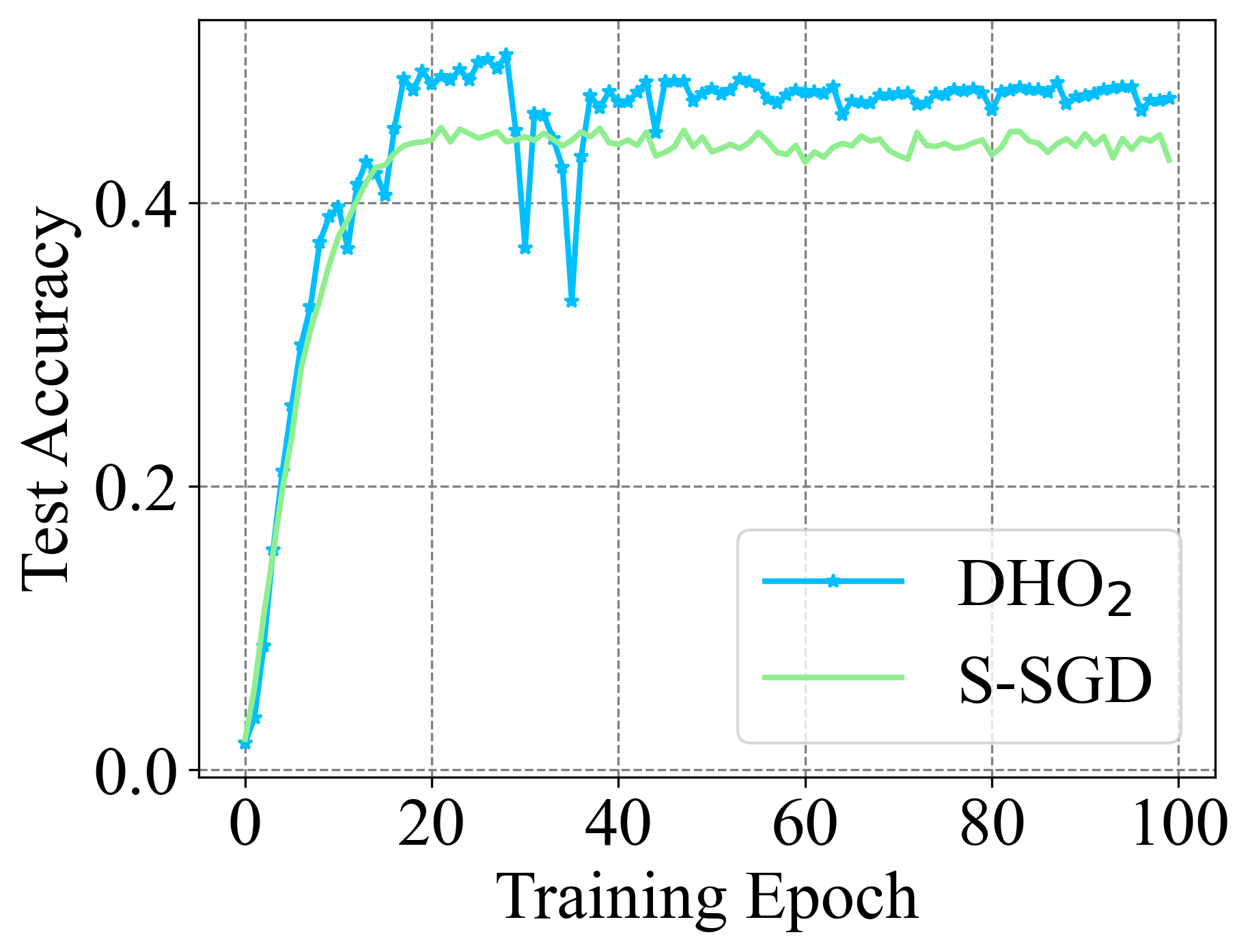}%
}
\caption{The test accuracy versus training epoch curves on the tiny-imagenet dataset, with different numbers of GPUs.}
\label{ex2}
\vspace{-0.3cm}
\end{figure}

\subsection{Scalability}
\begin{figure}[!tbp]
\centering
\subfloat[Time-to-Solution]{\includegraphics[width=4.4cm]{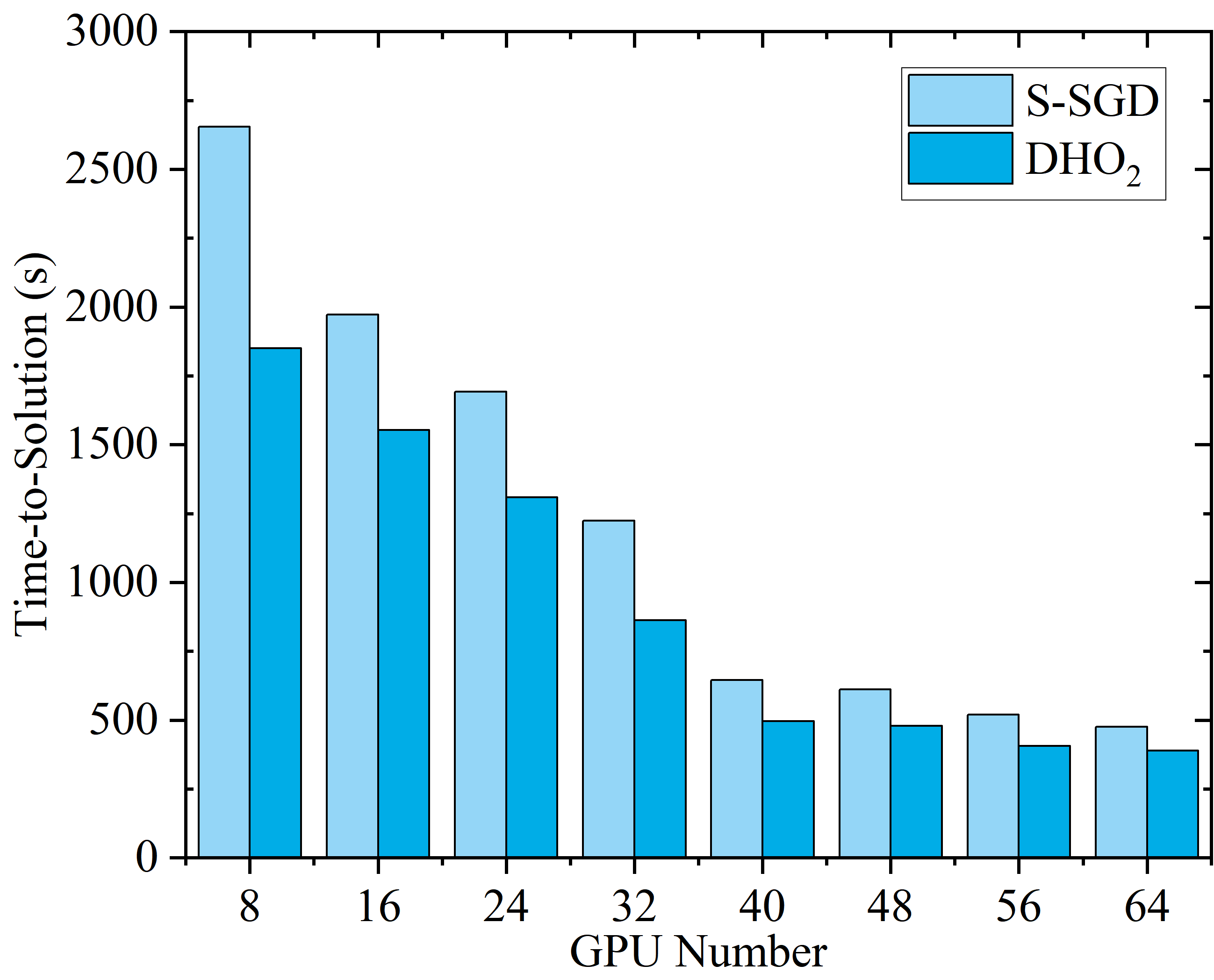}%
} 
% \hfil
\subfloat[Peak Memory Usage]{\includegraphics[width=4.4cm]{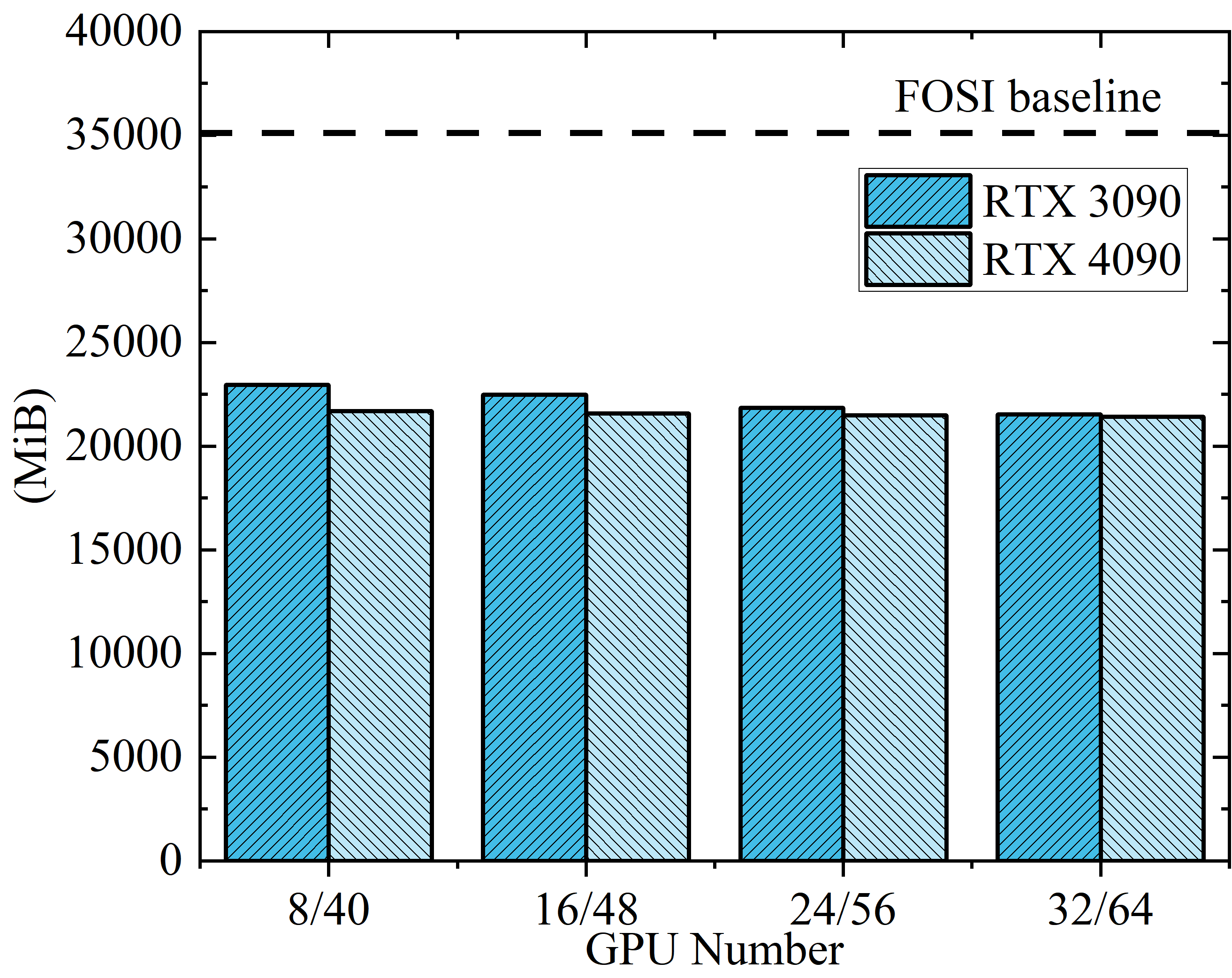}%
}
\caption{The time-to-solution and peak memory usage under different settings of the GPU number.}
\label{ex3}
\vspace{-0.3cm}
\end{figure}

We conduct our second experiment with different numbers of GPUS ($8\sim32$ 3090 GPUs and $40\sim64$ 4090 GPUs) to test the scalability of DHO$_2$. The batch size on each GPU is also fixed at 16. Since the global batch size is different when the GPU number varies, the optimal test accuracy also becomes different under the different settings. Therefore, the time-to-solution in the paper is regarded as the time to achieve the best accuracy within 100 training epochs. Although such a measure is unfair to DHO$_2$ since DHO$_2$ can converge to a better test accuracy than S-SGD in each setting of the GPU number (e.g. $53.32\%$ versus $49.27$\% on 24 GPUs), the experimental result can further prove the superiority of our framework. The learning rate is adjusted according to the global batch size, just as in \cite{krizhevsky2014one}. Fig. \ref{ex2} gives two examples of our scalability test. Meanwhile, Fig. \ref{ex3} demonstrates the time-to-solution and the peak memory usage under different settings of the GPU number. We can see from them that DHO$_2$ can bring approximately $4\%$ increase in the test accuracy, while it reduces $20\%\sim40\%$ time-to-solution, compared with S-SGD. We do think this acceleration is significant since the reduction brought by FOSI in our toy quick experiment is $11\%/16\%/12.5\%$ when the trained model is ResNet-18/50/101. Furthermore, Fig. \ref{ex3} shows that the peak memory usage of the simple distributed implementation of FOSI exceeds the maximum memory affordable to a single 3090/4090 GPU and our DHO$_2$ reduces $50\%$ memory usage and makes the distributed computing feasible. Last but not least, both the time-to-solution and peak memory usage demonstrate a sublinear relationship to the GPU number, which agrees with the complexity analysis of computation and memory described in Section \ref{p3p4}.

  \begin{figure}[!tbp]
\centering
\subfloat[swin-transformer-tiny]{\includegraphics[width=4.4cm]{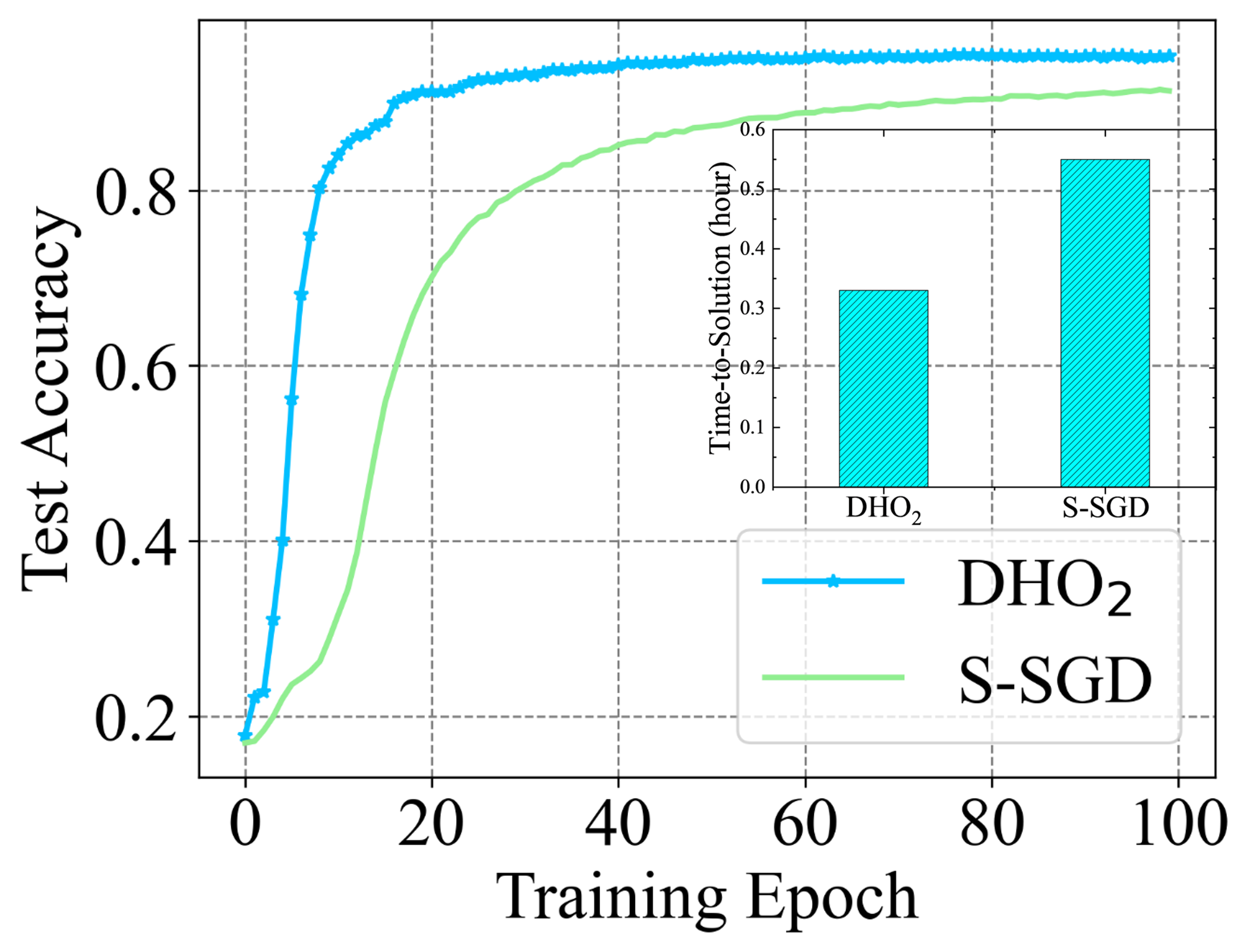}%
} 
% \hfil
\subfloat[swin-transformer-small]{\includegraphics[width=4.4cm]{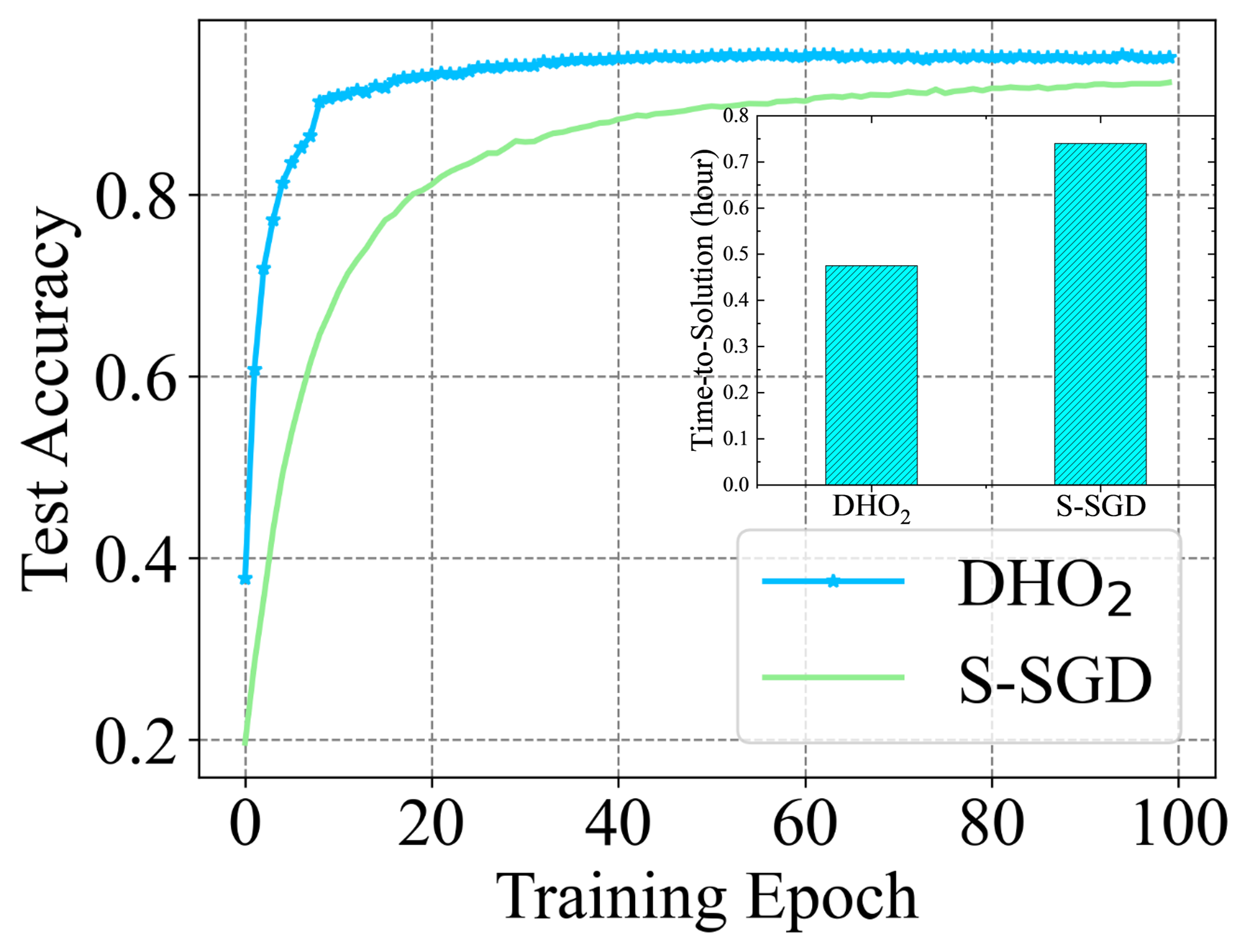}%
}
\caption{Extended study on the SVHN dataset with 32 3090 GPUs. We fine-tune a transformer-based model whose parameters are pre-trained on the ImageNet dataset and the base first-order optimizer is set as Adam with a learning rate of 1e-5. The batch size is 16 on each GPU.}
\label{extra}
\vspace{-0.3cm}
\end{figure}

\section{Conclusions}
In this paper, we have proposed a scalable FOSI-enabled distributed DNN training framework based on hybrid order optimization, namely DHO$_2$. Specifically, it incorporates a distributed Lanczos algorithm to balance the computation and memory cost for each GPU device, enabling the scalability of our framework. Then, an ADMM-like model update rule for the hybrid order optimization setting is designed to accelerate the distributed model training. Experimentally, our distributed Lanczos algorithm can reduce $50\%$ peak memory usage compared with the original one restricted on a single device. Meanwhile, the introduction of ADMM to the hybrid order optimization setting achieves up to $1.4\times\sim2.0\times$ speedup in the total training time and $4\%\sim5\%$ increase in the test accuracy, compared with other first-order and second-order frameworks.

% \section{Acknowledgments}
% This work is supported by National Natural Science Foundation of China under Grant No. 62302510.
% \vspace{0.1cm}
% \begin{figure*}[!tb]
% \centerline{\includegraphics[width=18cm]{fig_ex2.png}}
% \caption{The test accuracy versus training epoch curve when we train ResNet and Swin-Transformer ($K=G=24, l=0, I=200, \mathbb{O}=Adam$).}
% \label{ex2}
% % \vspace{-0.62cm}
% \vspace{-0.3cm}
% \end{figure*}
% \begin{figure*}[!tb]
% \centerline{\includegraphics[width=18cm]{fig_ex3.png}}
% \caption{The total training time and the training iteration time of our frameworks and other first-order optimizer frameworks.}
% \label{ex3}
% % \vspace{-0.62cm}
% \vspace{-0.3cm}
% \end{figure*}

\vspace{12pt}

\end{document}